\documentclass[a4paper,conference]{IEEEtran}
\ifCLASSINFOpdf
\else
\fi

\usepackage{subcaption}
\usepackage{graphicx}
\usepackage{algorithm}
\usepackage{algorithmic}
\usepackage[T1]{fontenc}

\begin{document}
%
% paper title
% Titles are generally capitalized except for words such as a, an, and, as,
% at, but, by, for, in, nor, of, on, or, the, to and up, which are usually
% not capitalized unless they are the first or last word of the title.
% Linebreaks \\ can be used within to get better formatting as desired.
% Do not put math or special symbols in the title.
\title{Pseudo Rehearsal using non photo-realistic images}

% author names and affiliations
% use a multiple column layout for up to three different
% affiliations
\author{\IEEEauthorblockN{Bhasker Sri Harsha Suri}
\IEEEauthorblockA{Computer Science Department\\
Indian Institute of Technology Tirupati\\
Tirupati, Andhra Pradesh\\
Email: cs18s506@iittp.ac.in}
\and
\IEEEauthorblockN{Kalidas Yeturu}
\IEEEauthorblockA{Computer Science Department\\
Indian Institute of Technology Tirupati\\
Tirupati, Andhra Pradesh\\
Email: ykalidas@iittp.ac.in}
}

% conference papers do not typically use \thanks and this command
% is locked out in conference mode. If really needed, such as for
% the acknowledgment of grants, issue a \IEEEoverridecommandlockouts
% after \documentclass

% for over three affiliations, or if they all won't fit within the width
% of the page, use this alternative format:
%
%\author{\IEEEauthorblockN{Michael Shell\IEEEauthorrefmark{1},
%Homer Simpson\IEEEauthorrefmark{2},
%James Kirk\IEEEauthorrefmark{3},
%Montgomery Scott\IEEEauthorrefmark{3} and
%Eldon Tyrell\IEEEauthorrefmark{4}}
%\IEEEauthorblockA{\IEEEauthorrefmark{1}School of Electrical and Computer Engineering\\
%Georgia Institute of Technology,
%Atlanta, Georgia 30332--0250\\ Email: see http://www.michaelshell.org/contact.html}
%\IEEEauthorblockA{\IEEEauthorrefmark{2}Twentieth Century Fox, Springfield, USA\\
%Email: homer@thesimpsons.com}
%\IEEEauthorblockA{\IEEEauthorrefmark{3}Starfleet Academy, San Francisco, California 96678-2391\\
%Telephone: (800) 555--1212, Fax: (888) 555--1212}
%\IEEEauthorblockA{\IEEEauthorrefmark{4}Tyrell Inc., 123 Replicant Street, Los Angeles, California 90210--4321}}

% use for special paper notices
%\IEEEspecialpapernotice{(Invited Paper)}

% make the title area
\maketitle

% As a general rule, do not put math, special symbols or citations
% in the abstract
\begin{abstract}
Deep Neural networks forget previously learnt tasks when they are faced with learning new tasks. This is called \textit{catastrophic forgetting}.
\textit{Rehearsing} the neural network with the training data of the previous task can protect the network from catastrophic forgetting.
Since rehearsing requires the storage of entire previous data, \textit{Pseudo rehearsal} was proposed, where samples belonging to the previous data are generated synthetically for rehearsal.
In an image classification setting, while current techniques try to generate synthetic data that is photo-realistic, we demonstrated that Neural networks can be rehearsed on data that is not photo-realistic and still achieve good retention of the previous task.
We also demonstrated that forgoing the constraint of having \textit{photo realism} in the generated data can result in a significant reduction in the consumption of computational and memory resources for pseudo rehearsal.
\end{abstract}

% no keywords

% For peer review papers, you can put extra information on the cover
% page as needed:
% \ifCLASSOPTIONpeerreview
% \begin{center} \bfseries EDICS Category: 3-BBND \end{center}
% \fi
%
% For peerreview papers, this IEEEtran command inserts a page break and
% creates the second title. It will be ignored for other modes.
\IEEEpeerreviewmaketitle

\section{Introduction}

Artificial neural networks have successfully demonstrated their ability to learn and perform on tasks that demand complex cognitive capabilities. 
 \cite{berner2019dota} \cite{rajpurkar2017chexnet}. However, current Neural Network architectures are incapable of learning new tasks sequentially. Whenever a neural network attempts to learn a new task, it forgets the task that it learnt previously. This problem is called \textit{catastrophic forgetting} \cite{french1999catastrophic}.
 
 In a classification setting, when a neural network is trained on a batch of data, the network learns a \textit{decision boundary} which it uses to classify the objects. However, when the same neural network is retrained with a new, independent batch of data, the previous \textit{decision boundary} gets distorted because of the weights updations. This distortion of the previous \textit{decision boundary} results in \textit{forgetting} of the previously learnt objects.
 
\subsection{Rehearsal and Pseudo-Rehearsal}

One simple way to prevent this forgetting is by retraining the neural network on training data of the previous task. 
The training data of the previous task is shuffled with the training data of the new task and is given to the neural network for training.
This \textit{interleaving} of the previous data with the new data will result in a decision boundary that is capable of handling both the tasks.
This technique is effective for many situations and is commonly referred to as \textit{Rehearsal}.
Though \textit{rehearsing} the Neural network on training data of the previous task is effective in mitigating \textit{catastrophic forgetting}, it is infeasible in a \textit{Continual Learning} setting where the neural network is required to learn new information throughout its lifetime.
In such scenarios, the memory requirements of the system would increase linearly with time, making the scheme undesirable.
Rehearsal can be generalised as:

\begin{equation}
    M_t = Train(B_1 \cup B_2 \cup B_3 \cup ..\cup B_t, M_{t-1})
\end{equation}

where $M_{t-1}$ is a Neural Network trained on some task $T_{t-1}$, $B_t$ is the training data for the task $T_t$ and $Train(B,M)$ is a function that trains the model $M$ on the data $B$. $P \cup Q$ represents a \textit{shuffled} union of the datasets $P$ and $Q$ and will be referred to as \textit{interleaving} from here on. 

To avoid storing the entire previous data, \textit{Pseudo Rehearsal} was proposed by \cite{robins1995catastrophic}. In \textit{Pseudo Rehearsal}, instead of storing the actual training data of the previous task, the training data is synthetically generated whenever the need arises. This synthetically generated training data of the previous task is \textit{interleaved} with the new task's training data and given to the neural network for training. \textit{Pseudo rehearsal} can be generalized as:

\begin{equation}
    M_t = Train(B_1^* \cup B_2^* \cup B_3^* \cup ..\cup B_t, M_{t-1})
\end{equation}
where,
\begin{equation}
    B_t^* = G(B_t)
\end{equation}

here $B_t^*$is the \textit{synthetic} version of $B_t$, generated using the \textit{ data generator} $G$.

% While all the previous approaches try to generate synthetic data that closely resemble the previous data, in the paper, we present a new approach where we instead try to generate training data that results in the same decision boundary for the neural network as the original data without worrying about the \textit{photo realism} of the generated samples.

\subsection{Generative Replay}
To generate the synthetic data, \cite{shin2017continual} proposed the usage of Generative Adversarial Networks (GAN)\cite{goodfellow2014generative} as our generator $G$.
The proposed system has a \textit{generator} network and a \textit{solver} network.
 Whenever the \textit{solver} is required to learn a new task, the \textit{generator} generates a synthetic version of the previous task's training data.
 This synthetic data is interleaved with the new task's training data and given to the \textit{solver} for training.
 In generative replay, the aim is to generate synthetic data that closely mimics the previous task's original data. 
 In an image classification setting, emphasis is given on generating \textit{photo realistic} synthetic samples for pseudo rehearsal.
 Even though GANs work well for \textit{visually simple} datasets like MNIST Handwritten digits\cite{krizhevsky2009learning}, their scalability to much more \textit{visually complex} datasets is questionable\cite{shin2017continual}\cite{parisi2019continual}. 
 Even though recent advancements in the field of Generative neural networks have demonstrated a significant increase in the photo-realism of the 
  \begin{figure*}
\begin{center}
\begin{subfigure}{.3\textwidth}
    \centering
    \includegraphics[width=5.5cm]{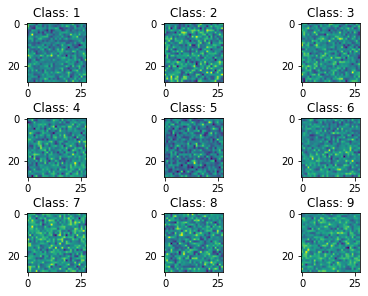}
    \caption{Our technique}
    \label{fig:genetic_images}
\end{subfigure}
\begin{subfigure}{.3\textwidth}
    \centering
    \includegraphics[width=5.5cm]{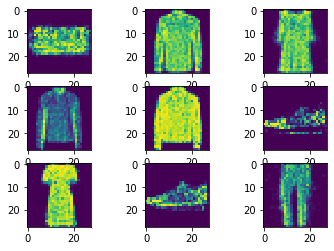}
    \caption{GAN generated images}
    \label{fig:gan_images}
\end{subfigure}
\begin{subfigure}{.3\textwidth}
    \centering
    \includegraphics[width=5.5cm]{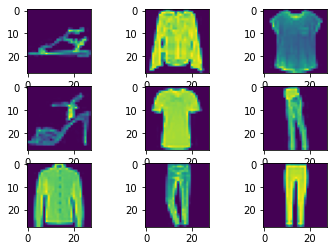}
    \caption{Original images}
    \label{fig:original_images}
\end{subfigure}
\caption{A comparison of Genetically generated images and GAN generated images with Original images.}
\end{center}
\end{figure*}

 generated synthetic images\cite{atkinson2018pseudo}\cite{karras2019style}, they require significant computational resources and careful monitoring by a human, which make them impractical in a \textit{continual learning} scenario.
 Apart from generating \textit{photo-realistic} images, the GANs are required to generate synthetic images that can capture the wide \textit{visual diversity} of images in the original dataset. Only then can the generated synthetic dataset protect the \textit{decision boundary} from significant distortions on pseudo rehearsal. 
 In addition to the above-mentioned disadvantages, another disadvantage of using GANs for Generative replay is the need to maintain a separate neural structure that demands additional computational and memory resources.

In this paper, we demonstrate that generating synthetic data that is spatially close to the original data and which closely envelops the initial \textit{decision boundary} in the latent space is enough to serve the purpose of \textit{pseudo rehearsal} without worrying about \textit{photo realism} of the generated samples. Instead of using a separate neural structure to generate the synthetic data, we are generating the data from the \textit{solver} neural network itself. We demonstrated our proposed technique on MNIST Handwritten digits and also on MNIST Fashion dataset \cite{DBLP:journals/corr/abs-1708-07747} which is much more \textit{visually complex} compared to MNIST Handwritten digits. It was also experimentally shown that the proposed technique requires significantly lesser time to generate synthetic data while consuming modest memory and computational resources.

Even though we demonstrated our technique in an image classification setting, the proposed technique applies to non-image scenarios as well.

\section{Proposed Method}

Instead of trying to generate \textit{photo-realistic} images, we propose to generate images, which when trained upon, will result is similar \textit{decision boundary} as the original data. If we can generate synthetic data for pseudo rehearsal, which can protect the decision boundary from significant distortions, then we can effectively forego the constraint of trying to generate \textit{photo-realistic} synthetic images.

Robins\cite{robins1995catastrophic} was the first to explore the idea of using randomly generated vectors for \textit{pseudo rehearsal}.
However, randomly generating vectors leads to generating synthetic data with class imbalance problems. 
The generated images might be unevenly distributed in the latent space which might result in a distorted decision boundary when trained upon.
In our proposed technique, we address the class imbalance problem using Genetic Algorithms.
We first generate synthetic samples that are spatially close to the original data in the latent space.
We then subject the generated data to an \textit{enrichment phase} where we ensure that the generated synthetic data is enveloping the initial \textit{decision boundary} while remaining spatially close to the original data.
The resultant synthetic dataset looks like random images (Figure \ref{fig:genetic_images}), however, it has the capability of protecting the decision boundary of the neural network when used for pseudo rehearsal.

\begin{figure*}
\begin{subfigure}{.5\textwidth}
    \centering
    \includegraphics[width=5cm]{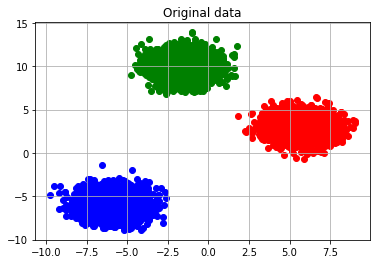}
    \caption{Original dataset}
    \label{fig:orginal}
\end{subfigure}
\begin{subfigure}{.5\textwidth}
    \centering
    \includegraphics[width=5cm]{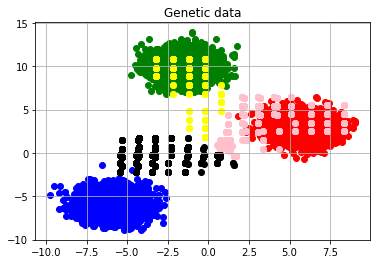}
    \caption{Genetically evolved datapoints}
    \label{fig:genetic}
\end{subfigure}
\begin{subfigure}{.5\textwidth}
    \centering
    \includegraphics[width=5cm]{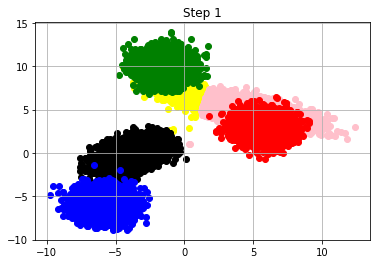}
    \caption{First step of enrichment}
    \label{fig:step1}
\end{subfigure}
\begin{subfigure}{.5\textwidth}
    \centering
    \includegraphics[width=5cm]{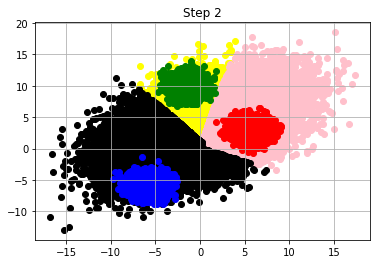}
    \caption{Second step of enrichment}
    \label{fig:step2}
\end{subfigure}
\caption {Synthetic data generated using the proposed method. Figure (\ref{fig:orginal}) image shows the original dataset. Figure (\ref{fig:genetic}) shows the genetically generated images. Figure (\ref{fig:step1}) and Figure (\ref{fig:step2}) show the enrichment process where the number of samples and the spread of the synthetically generated samples are increased. The points in red, green and blue are the original data points while the points in pink, yellow and black are synthetically generated data points.}
\label{fig:Blobs}
\end{figure*}

\subsection{Genetic Pseudo Rehearsal}
To generate the synthetic data $B_T^*$ for some task $T$, we use \textit{Genetic algorithms} as our generator $G$ and our \textit{solver} Neural network $N_{T}$ as our fitness function for the Genetic Algorithm. Here, $N_{T}$ is the neural network that was trained on the task $T$. To generate $B_T^*$, the Genetic Algorithm begins with a random population ($P$). Next, the fitness of each sample in $P$ is determined by $N_{T}$ by giving them the softmax confidence of belonging to the task $T$. For example, when we want to generate synthetic data for MNIST Handwritten digits, we first take a Neural Network $N_H$ which was already trained to recognize MNIST Handwritten digits. Then, a random population of images is generated and the softmax confidence of belonging to class 0 is given by $N_H$ to all the samples in the population. This softmax confidence of belonging to class 0 is considered as the \textit{fitness} of the sample. The fittest 25\% of the population are chosen based on their fitness scores and propagated to the next generation. A series of mutation and cross-over operations are performed on the fittest 25\% to create the next generation. This is repeated until a satisfactory fitness $\tau$ is achieved by all the samples of a population in a generation. This whole process is then repeated for all the classes from 0 to 9.

After generating this initial population, an \textit{Enrichment phase} follows, where the goal is to increase the number of samples in the generated population and envelop the \textit{decision boundary} in the latent space. 
In the first step of enrichment, each class is assumed as a Gaussian probability distribution(Equation \ref{eq:gaussian}) and the mean($\mu$) and covariance matrix($\Sigma$) of all the samples in the class is found. 
\begin{equation}
    g(x_1...x_k) = \frac{1}{\sqrt{(2\pi)^k|\Sigma|}}\exp(-\frac{1}{2}(x-\mu)^{T}\Sigma^{-1}(x-\mu))
    \label{eq:gaussian}
\end{equation}

To increase the number of samples per class, synthetic points are sampled using the mean($\mu$) and covariance matrix($\Sigma$) of the Gaussian. 
In the second step, one single Gaussian is fitted to the entire synthetic data and more synthetic samples are generated.
The data that is obtained after the second step is our synthetic data.
To get a visual perspective of the process, please refer to Figure \ref{fig:Blobs}.
The proposed technique was used to generate synthetic data for the Blobs dataset which was created using \textit{make blobs} function in \textit{sklearn} package.
The Gaussians were implemented using the \textit{Gaussian Mixture Models} function in \textit{sklearn}.

As Figure \ref{fig:Blobs} suggests, the initial population that was created using Genetic Algorithms is not sufficient and cannot be considered as a suitable replacement of original data for \textit{pseudo rehearsal}. The number of samples that were generated is not large enough and the samples are not enveloping the decision boundary. Both these issues are addressed in the \textit{Enrichment phase}. In this first step, we are fitting a Gaussian to each class and generating synthetic points. This results in an increase in the number of points per class as well as \textit{scatter} of points. In the second step, since one single Gaussian is fitted to the entire dataset, the points that are now generated are scattered evenly from the mean point. This creates a smooth sheet of points that cover the decision boundary quite effectively. This data can now be used as a suitable replacement for the actual data for \textit{pseudo rehearsal}.

The algorithm has been described in Algorithm \ref{Algo}.
To increase the diversity of the population and allow for exploration of the latent space, the concept of \textit{cultures} was explored.
To maintain the diversity of population, multiple populations were initialized and evolved independently from each other.
Each of these independent population is called a \textit{culture}.
Different \textit{selection mechanisms} for choosing the fittest individuals in a given generation were also studied.
Their effect on the generated population was also explored and presented in the Ablation studies section. 

\begin{algorithm}[tb]
   \caption{Synthetic data generation for 1 class}
   \label{Algo}
\begin{algorithmic}
  \STATE {\bfseries P:} \{$x_0, x_1, x_2 ... x_m$\} // Random population
  \STATE {\bfseries $|P|$} = $m$
    \WHILE{$min(f(x)) | \forall x \in P \leq \tau$} 
        \STATE $P^{'}$ = $\{<x,f(x_t)>\} | \forall x \in P$
        
        \STATE $P^{D} = P^{'}$ in descending order of $f(x) , \forall x \in P^{'}$
        \STATE $P^{*} =P^{D}[i]$ where $i \in [0,m*0.25]$
        \STATE $C$ = crossover$(P^{*}[j],P^{*}[j+1])$ where $j$ in $[0,|P^{*}|]$
        
        \STATE $M$ = mutation$(x) | \forall x \in P^*$
        \STATE $M_C$ = crossover$(X^m[j],X^m[j+1])$
        \STATE //where $j \in [0,|P^*|]$
        \STATE $P_{new} = P^* \cup C \cup M \cup M_C$
        \STATE $P = P_{new}$
    \ENDWHILE
    \STATE 
    \STATE Here $f(x_t)$ = $\frac{e^{z_t}}{\sum_{j=1}^K e^{z_j}}$
    \STATE where $t$ is the target class
    \STATE       $z$ is the softmax output of the sample
    \STATE       $K$ is the total number of classes
\end{algorithmic}
\end{algorithm}

\begin{figure*}
\begin{subfigure}{.5\textwidth}
    \centering
    \includegraphics[width=8cm]{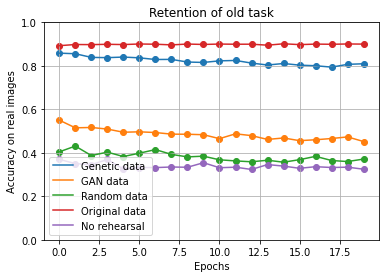}
    \caption{Retention of old task}
    \label{fig:retention_experiment}
\end{subfigure}
\begin{subfigure}{.5\textwidth}
    \centering
    \includegraphics[width=8cm]{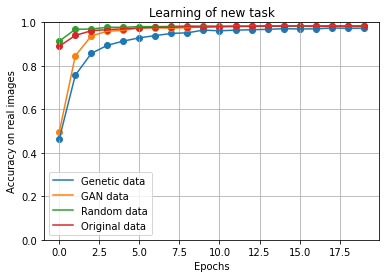}
    \caption{Learning of new task}
    \label{fig:learning_exp1}
\end{subfigure}
\begin{subfigure}{.5\textwidth}
    \centering
    \includegraphics[width=8cm]{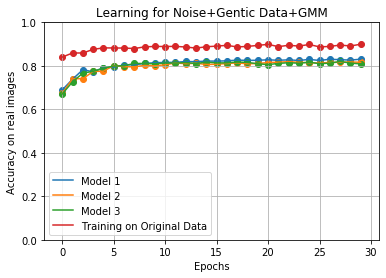}
    \caption{Demonstration of classification abilities on real images when trained on synthetic images}
    \label{fig:genetic_learning}
\end{subfigure}
\begin{subfigure}{.5\textwidth}
    \centering
    \includegraphics[width=8cm]{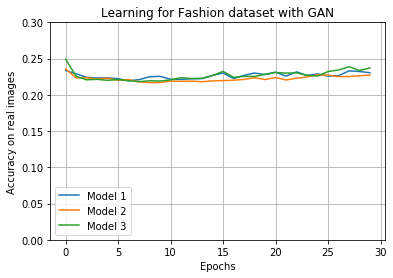}
    \caption{Classification abilities when the neural networks are trained on GAN generated images}
    \label{fig:gan_learning}
\end{subfigure}
\caption {The learning and retention behaviour of neural network under various rehearsal and pseudo-rehearsal schemes}
\label{fig:Benchmarking}
\end{figure*}

\section{Experiments}
Four different experiments have been conducted to investigate the quality of the synthetic data that was generated using the proposed algorithm.
In the first experiment, a neural network is required to learn MNIST Fashion and MNIST Handwritten digits datasets serially. The neural network is allowed to \textit{pseudo rehearse} on MNIST Fashion products while learning MNIST Handwritten digits. The retention capacity of the neural network using our technique was benchmarked against other techniques in the literature. 
In the second experiment, we aim to qualitatively compare the synthetic data with the original data by training neural networks on synthetic data and measuring their accuracy on real images.
In the third experiment, we try to quantify the difference between a neural network that was trained on original data versus a neural network that was trained on synthetic data by proposing a new metric called \textit{Agreement score}.

\subsection{Interleaved pseudo rehearsal}

In this experiment, the retention behaviour of the neural network under various pseudo rehearsal schemes was investigated.
The neural network is required to learn two tasks sequentially.
While the network is learning the second task, the synthetically generated data of the first task is interleaved with the training data of the second task and is presented to the neural network for training.
The accuracy of the model on the previous task while learning the new task was recorded and plotted.
For this experiment, the Neural network was first trained on MNIST Fashion products.
Next, synthetic data for MNIST Fashion products was generated using various schemes and interleaved with training data of MNIST Handwritten digits.
This data was presented to the neural network as training data for 30 epochs.
The accuracy of the network on MNIST Fashion products was measured at each epoch.
The X-axis represents the epoch number and the Y axis represents the accuracy on the previous task.
A standard multi layer perceptron was used for this experiment. The network had 1 hidden layer with \textit{relu} activation function. Binary cross entrophy loss was used with \textit{adam} optimiser for the network.
The model's behaviour was tested for five cases. In the first case, the original training data of MNIST Fashion was interleaved with the data of Handwritten digits and was given to the network for training.
In the second case, the synthetic data generated using a Generative adversarial network was interleaved with the Handwritten digits.
In the third case, synthetic data generated using our proposed technique was interleaved with the Handwritten digits.
In the fourth case, randomly generated vectors were mixed with Handwritten digits and given to the model.
The fifth case was the \textit{control case} where the neural network was not allowed to rehearse.
\begin{figure*}
\begin{subfigure}{.3\textwidth}
    \centering
    \includegraphics[width=5.5cm]{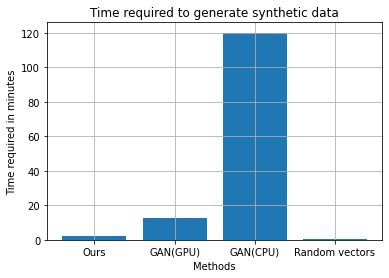}
    \caption{Time required by all the methods}
    \label{fig:time_consumed}
\end{subfigure}
\begin{subfigure}{.3\textwidth}
    \centering
    \includegraphics[width=5.5cm]{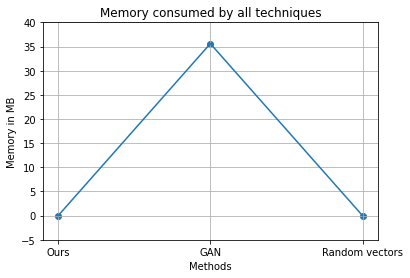}
    \caption{Memory consumed by all the methods}
    \label{fig:memory_consumed}
\end{subfigure}
% \begin{figure*}[hbt!]
\begin{subfigure}{.3\textwidth}
\includegraphics[width=5.5cm]{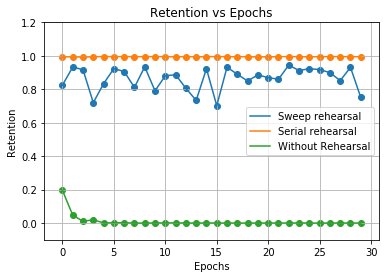}
\caption{Comparison of various rehearsal schemes}
\label{fig:interleave}
\end{subfigure}

\begin{center}
\begin{subfigure}{.4\textwidth}
\includegraphics[width=6cm]{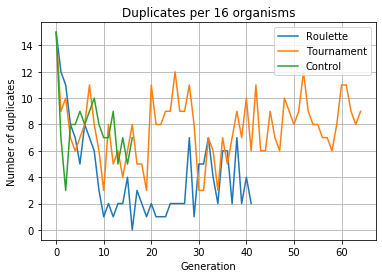}
\caption{Duplicate organisms for various selection schemes}
\label{fig:selection}
\end{subfigure}
% \end{figure*}
\begin{subfigure}{.4\textwidth}
\includegraphics[width=6cm]{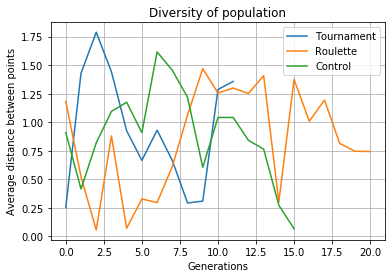}
\caption{Distance between organisms for various schemes}
\label{fig:distance_diversity}
\end{subfigure}
\end{center}

\caption{Results of Ablation studies and computational resource consumption comparisons}
\end{figure*}

The results have been plotted in Figure \ref{fig:retention_experiment}. As the results suggest, the highest retention capacities were observed for rehearsing on original dataset.
Synthetic data that was generated using our proposed technique also demonstrated near-perfect retention capacities of MNIST Fashion products.
The experiment clearly demonstrates the short comings of using \textit{photo realistic} data for pseudo rehearsal. Since the Generative Adversarial Network was unable to generate synthetic images of fashion products that were both \textit{photo realistic} and \textit{visually diverse}, the overall retention capacity of the neural network when trained on that synthetic data is relatively low.
Pseudo rehearsal on random vectors showed almost similar retention capacities as that of performing no rehearsal at all.
This experiment highlights the fact that the proposed method is scalable to datasets that are \textit{visually complex}. While generative techniques struggle to generate photo realistic synthetic samples, the proposed technique allows for higher retention capacities by forgoing the \textit{photo realism} constraint.
Figure \ref{fig:learning_exp1} shows the accuracy of all the techniques on MNIST Handwritten digits during the experiment. 

\subsection{Can we train models on synthetic data?}
In this experiment we further try to justify that the synthetic data generated using our method is a suitable replacement for the original data for pseudo rehearsal.
One simple way to test the equality of synthetic and original datasets is by asking the question: \textit{What happens if we train a random neural network on these synthetically generated images?}
Can a network that is trained on synthetic data classify real images?
If a network that is trained on synthetic images can successfully classify real images, then we can confirm that the synthetic data has all the properties of the original dataset and can be considered as a suitable replacement for it.
To answer this question, three randomly initialized neural networks with one hidden layer with \textit{relu} activation were taken. The synthetically generated data was provided as training data and their accuracy on 10,000 real images was tested. 
The experiment was performed for MNIST Fashion products dataset where the neural network was trained on synthetic version of the Fashion dataset and was tested on real images of Fashion products.
The proposed technique was again benchmarked against GAN-generated synthetic data. The results have been plotted in the Figure \ref{fig:genetic_learning} and Figure \ref{fig:gan_learning}.
It is surprising to find that neural networks that were trained on images that look like random noise (Figure \ref{fig:genetic_images}) successfully classify real life images (Figure \ref{fig:original_images}).
As the results suggest, the neural networks that were trained on synthetic data that was generated using our proposed technique reached a maximum accuracy of 81\% by the end of training. 
In comparison, the neural networks that were trained on original data reached a peak accuracy of 91\% while the neural networks that were trained on synthetic data generated by GANs only reached a peak accuracy of 24\% at the end of training. 
Even though GAN was able to generate photo-realistic synthetic samples of fashion products (Figure \ref{fig:gan_images}), it failed to generate images that were \textit{visually diverse}. For example, a GAN can easily generate photo realistic images of a dress, however, automatically generating images of dresses with diverse shapes, colors and patterns that are possible in the real world is a difficult task.
Refer to Figure \ref{fig:gan_images} for GAN generated images.

\subsection{Agreement score}

In this paper, we propose a new metric called the \textit{Agreement score}.
The main purpose of this metric is to quantify the difference of predictions between a model that has been trained on original data and a model that has been trained on synthetic data.
While other metrics like accuracy compare the total number of correct predictions, we propose to compare all the predictions between a model trained on original data and a model trained on synthetic data.
In the proposed metric, the predictions of both the models are compared on one standard test set.
Here, we count the total number of matching predictions between the two models on a common test set, rather than counting correct predictions alone.
If the \textit{Agreement score} is high, then it can be implied that the model that is trained on the synthetic data has a behavior that is similar to the model that was trained on the original data.
The agreement score can be calculated using the formula
\begin{equation}
    \alpha(P_M,P_N) = \frac{\theta}{|T|} * 100
\end{equation}

where, $P_M$ and $P_N$ are the predictions of model $M$ and model $N$ on some test dataset. $\theta$ is the number of identical predictions, $|T|$ is the size of the test data.

In this experiment, one neural network was initially trained on the MNIST Fashion dataset with 60,000 train samples from the original data. Then two similar neural networks that were initialized with the same initial weights of the first network were trained on Genetically generated and GAN generated data. All the three neural networks were made to make predictions on a constant test set of 10,000 images from the original dataset. First, the \textit{Agreement score} between the model trained on original data and model trained on Genetic data was calculated. Then the \textit{Agreement score} for a model trained on original data and model trained on GAN generated data was calculated. The experiment was repeated for MNIST Handwritten digits dataset as well. The results have been tabulated in the below table.

.
\begin{center}
    \begin{tabular}{|c|c|c|}
    \hline
        Dataset & GAN Data & Ours  \\
        \hline \hline
        MNIST Handwritten digits & \textbf{95.346\%} & 83.675\% \\
         \hline
         MNIST Fashion products &  52.459\% & \textbf{80.977\%} \\
         \hline
    \end{tabular}
    % \caption {Results of Agreement score experiment}
    \label{tab:agreement_score}
\end{center}

As the results suggest, Generative Adversarial networks are quite competent at generating synthetic data for visually simple datasets like MNIST Handwritten digits.
However, synthetic data generated by them for visually complex datasets like MNIST Fashion is not suitable for pseudo rehearsal.
The Agreement score drops drastically when the neural network is trained on synthetic images of Fashion products that were generated by a Generative Adversarial network.
This experiment demonstrates that the proposed technique is scalable to visually complex datasets as well.

\subsection{Comparison of time required}
The time consumed by both the techniques to generate synthetic data has been recorded and graphed. While the Generative replay technique required 12.6 minutes to generate the synthetic data, our proposed technique required only 2.4 minutes. It also has to be noted that the GAN was running on Tesla P100 GPU while the proposed technique was only using Intel Xeon Dual Core 2.5 GHz processor. To make the comparison fair, the GAN was also run on the Intel processor without GPU support and the results have been plotted in Figure \ref{fig:time_consumed}.
The GAN consumed approximately 120 minutes to run on Intel Xeon Dual Core CPU. While the generation of random vectors required the least amount of time, the poor performance of the technique prohibits its adoption. Since genetic algorithms can utilize GPU architectures because of their parallelizable nature, much more reduction in time can be expected. Implementing our proposed algorithm on GPU architecture will be a part of our future work.

\subsection{Memory consumption comparison}
In a continual learning scenario, it is highly essential to observe the additional memory resources that are consumed by the pseudo rehearsal scheme.
When we compare the additional memory that is required by all the pseudo rehearsal schemes we can argue that both random vector rehearsal and our proposed technique do not require any permanent short term memory to store any kind of additional neural structures.
All the synthetic data is generated on demand using the \textit{solver} network itself.
However, when we use a Generative adversarial network generating the synthetic data it has to be noted that storing of the GAN requires additional memory resources.
The additional memory requirements that were observed during our experimentation have been plotted in Figure \ref{fig:memory_consumed}.

\begin{figure*}
\begin{subfigure}{.3\textwidth}
    \centering
    \includegraphics[width=5cm]{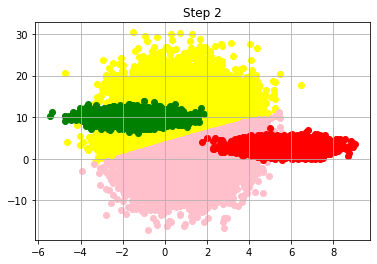}
    \caption{Synthetic data}
    \label{fig:blobs_synthetic_data}
\end{subfigure}
\begin{subfigure}{.3\textwidth}
    \centering
    \includegraphics[width=5cm]{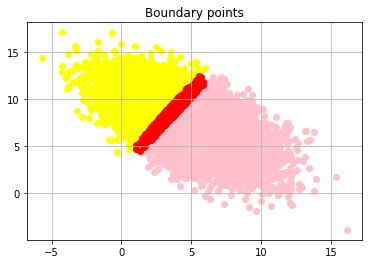}
    \caption{Boundary points using SVM method}
    \label{fig:svm_boundary_points}
\end{subfigure}
\begin{subfigure}{.3\textwidth}
    \centering
    \includegraphics[width=5cm]{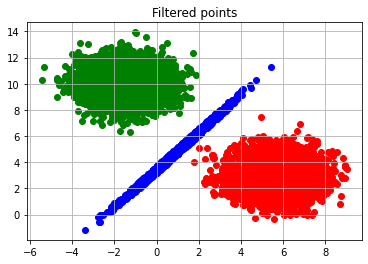}
    \caption{Boundary points using confidence score}
    \label{fig:neural_boundary_points}
\end{subfigure}
% \end{figure*}
% \begin{figure*}
\begin{center}

\begin{subfigure}{.3\textwidth}
    \centering
    \includegraphics[width=5cm]{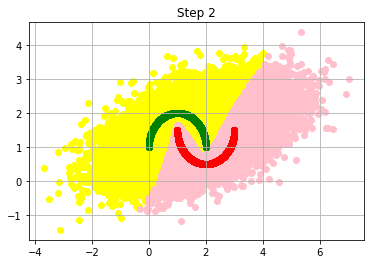}
    \caption{Generated synthetic data}
    \label{fig:make_moons}
\end{subfigure}
\begin{subfigure}{.3\textwidth}
    \centering
    \includegraphics[width=5cm]{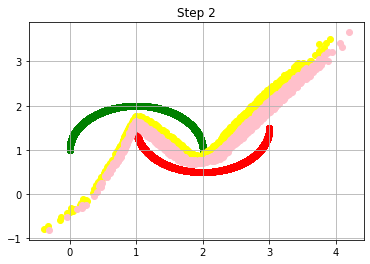}
    \caption{Boundary points found using softmax confidence}
    \label{fig:neural_boundary_points_2}
\end{subfigure}
\end{center}
\caption{Boundary point filtering on synthetic datasets}
\end{figure*}

\section {Ablation Studies}

There are many schemes to select the fittest population in genetic algorithms. In our ablation studies, we systematically evaluated various selection mechanisms for selecting the fittest organisms in a generation. In addition to that, we also investigated various ways in which we could \textit{interleave} synthetic data with original data. The methods and results of our investigations are given below.
\subsubsection {Comparision of selection techniques}
To select the fittest 25\% of the population, 3 different selection mechanisms were explored: \textit{Roulette wheel selection}, \textit{Tournament selection}, \textit{Linear selection}. 
In \textit{Roulette wheel selection}, each sample in the population has a probability of getting selected that is directly proportional to the fitness of the respective organism, which in our case was the softmax confidence of the target class.
In \textit{Tournament selection}, $p$ percent of the population is randomly chosen for extinction first and then the fittest 25\% samples are selected from the remaining population. 
In Linear selection, the fittest 25\% are chosen directly based on their fitness score.

Two metrics were considered to quantify the fitness of the generated population for a given selection mechanism.
The first was the average Euclidean distance between all the organisms in a given generation.
This gave us the \textit{spread} of the generated population in the latent space.
The second is the \textit{number of duplicates} generated in each generation.
The results are shown in Figure \ref{fig:selection} and Figure \ref{fig:distance_diversity}. The experiments suggest that there is no significant difference between the diversity of the populations generated using the 3 selection mechanisms.
Hence, the \textit{linear selection} mechanism was chosen because of its simplicity.

\subsubsection {Comparison of rehearsal schemes}
There are different techniques to interleave the old and new data.
The results of investigating two of the most common techniques have been presented here.
In this work, the \textit{interleaving} of the synthetic data was done by training the neural network on synthetic data of the previous task for one epoch and then training on the new dataset in the next epoch.
This alternating scheme of training the synthetic old data and the new data was done for 30 cycles.
This scheme of rehearsal was named as \textit{Serial rehearsal}. 
We compared \textit{Serial rehearsal} with Sweep rehearsal which was proposed by Robins et al\cite{robins1995catastrophic}.
We found out that, simple alternate training of old and new datasets for one epoch each was enough to demonstrate significant retention capacities.
The results have been plotted in Figure \ref{fig:interleave}.

\section{Discussion}
In this work, we tried to demonstrate that neural networks can be trained on synthetic data that are not photo-realistic and still show high retention capacities on pseudo rehearsal.
However, in the current method, very large number of synthetic samples had to be generated in the enrichment phase to ensure that the generated synthetic data envelops the decision boundary tightly.
In all the experiments, the synthetic data had 1 million synthetic samples in it.
It has to be noted that, to keep the comparison fair, we generated 1 million synthetic samples for Generative replay and random vector rehearsal as well. 
Our future work will focus on generating synthetic data that are much closer to the decision boundary. 
This will drastically reduce the total number of samples needed to sufficiently envelop the decision boundary.
Few initial steps have been taken in this direction and we succeeded in filtering out synthetic datapoints that are closer to the decision boundary.
In our first appraoch, the synthetic data that was generated after the enrichment phase was given as training data to a Support Vector Machine\cite{cortes1995support}.
The support vectors of the SVM would 
\begin{figure*}
\begin{subfigure}{.5\textwidth}
    \centering
    \includegraphics[width=8cm]{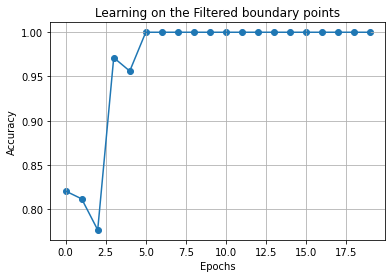}
    \caption{Learning of neural network on synthetic data of make moons dataset}
    \label{fig:Learning_make_moons}
\end{subfigure}
\begin{subfigure}{.5\textwidth}
    \centering
    \includegraphics[width=8cm]{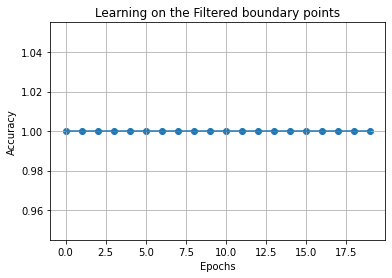}
    \caption{Learning of neural network on synthetic data of make blobs dataset}
    \label{fig:Learning_make_blobs}
\end{subfigure}
\caption{Learning behavior of neural networks on boundary points}
\end{figure*}

be our desired points as they would be the points that are closest to the decision boundary.
The technique was tried on synthetic datasets like \textit{makeblobs} and the results are shown in the Figure \ref{fig:svm_boundary_points}.
In the second approach, the \textit{solver} neural network itself was used to filter the synthetic data for points closer to the decision boundary.
This was done by finding the points that had least standard deviation in their softmax confidence list.
This technique was also demonstrated on \textit{make blobs} and \textit{make moons} (Figure \ref{fig:make_moons}) datasets of \textit{sklearn} package.
The results are shown in Figure \ref{fig:neural_boundary_points} and Figure \ref{fig:neural_boundary_points_2}.
To test the quality of the generated data, we trained two neural networks on these boundary points and tested on original data. 
The results have been plotted in Figure \ref{fig:Learning_make_moons} and Figure \ref{fig:Learning_make_blobs}.
As the results suggests, when neural networks were trained on these boundary points, they achieved 100\% accuracy in less than 5 epochs.
Our future work will focus on reducing the total number of required points using one of the above mentioned techniques.
Efforts to implement the current algorithm on GPU architecture will also be made.
% The code is being made available at the following URL to promote reproducibility of our research:

% https://github.com/BhaskerSriHarsha/Genetic-Pseudo-Rehearsal

\section{Conclusion}
In this work, we demonstrated that in an image classification setting, pseudo rehearsal can be performed by ignoring the\textit{ photo-realism} of the generated samples.
We also showed that by ignoring the constraint of photo-realism of the generated synthetic samples, we can achieve high retention capacities of the previous task while consuming modest computational and memory resources.
We also demonstrated that the proposed technique is scalable to \textit{visually complex} datasets unlike existing techniques in the literature.

\bibliographystyle{ieeetr}
\bibliography{IEEEfull}

% that's all folks
\end{document}